%% file: iclr2026_conference.tex
\documentclass{article} % For LaTeX2e
\usepackage{iclr2026_conference,times}

\input{math_commands.tex}

\usepackage{hyperref}
\usepackage{url}
\usepackage{graphicx}
\usepackage{subcaption}
\usepackage{booktabs}
\usepackage{xcolor}

\title{LookSharp:\,Attention Entropy \\ Minimization for Test-Time Adaptation}

\author{Yash Mali$^{1}$, Evan Shelhamer$^{1, 2}$\\
  University of British Columbia$^{1}$, Vector Institute$^{2}$\\
  \texttt{ymali@mail.ubc.ca}
}

\iclrfinalcopy % Uncomment for camera-ready version, but NOT for submission.

\begin{document}

\maketitle

% \todo{Name recommendations would be welcome! LookHarder, LookSharper, LookMore, FocusTTA, "DARTT: Diffuse-Attention Reduction at Test-Time", ... :)}

\begin{abstract}
Test-time adaptation (TTA) updates models during inference to reduce error on distribution shifts. While entropy minimization over the output distribution has proven effective as a TTA loss, we study using the intermediate distributions computed by transformers in the attention mechanism. We propose \textit{LookSharp}, which minimizes the entropy of CLS-to-patch attention in the final layer as a novel TTA objective, encouraging the model to maintain focused attention on shifted data. We demonstrate that attention entropy minimization improves robustness on ImageNet-C \citep{hendrycks2019robustness}. We also show that it is complementary to output entropy minimization and maintains performance on clean data.
\end{abstract}

\section{Introduction and Related Work}
Deep networks achieve impressive performance on in-distribution data but often fail catastrophically when deployed on data from shifted distributions \citep{hendrycks2019robustness}.
Recent TTA methods have explored entropy minimization over the output distribution, which encourages the model to make confident predictions at test time.
While effective, this approach treats the feature extractor as a black box and ignores internal representations that could guide adaptation.
Vision Transformers (ViTs) \citep{dosovitskiy2021imageworth16x16words}, which have become the dominant architecture for visual recognition due to their scalability, offer attention distributions over image patches that explicitly capture spatial relationships and feature importance \citep{fullerlookwhere}.

\begin{figure}[htbp]
\centering
\begin{subfigure}[b]{0.48\textwidth}
  \centering
  \includegraphics[width=\textwidth]{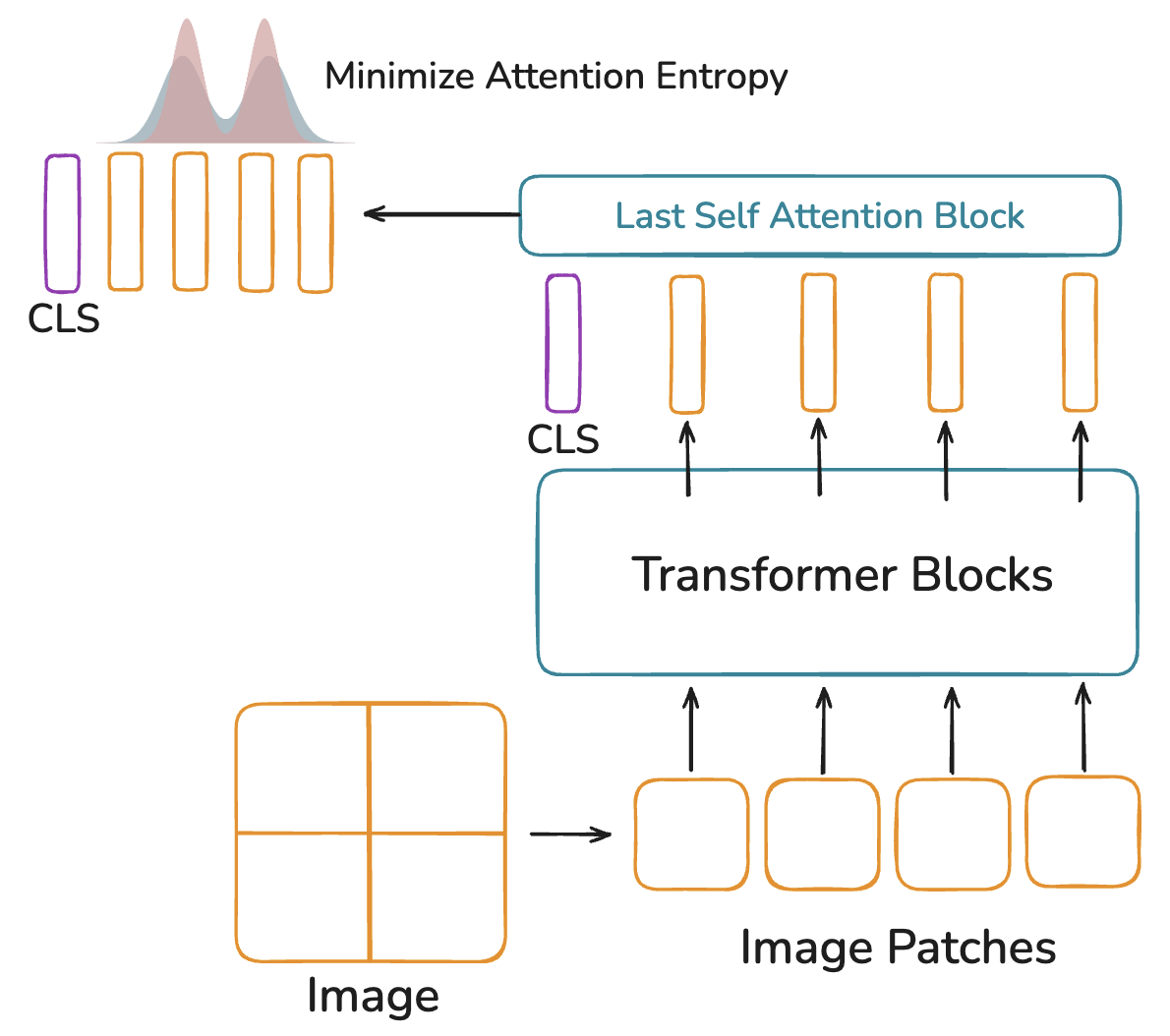}
  \caption{Summary of the Method}
  \label{fig:summary}
\end{subfigure}
\hfill
\begin{subfigure}[b]{0.48\textwidth}
  \centering
  \includegraphics[width=\textwidth]{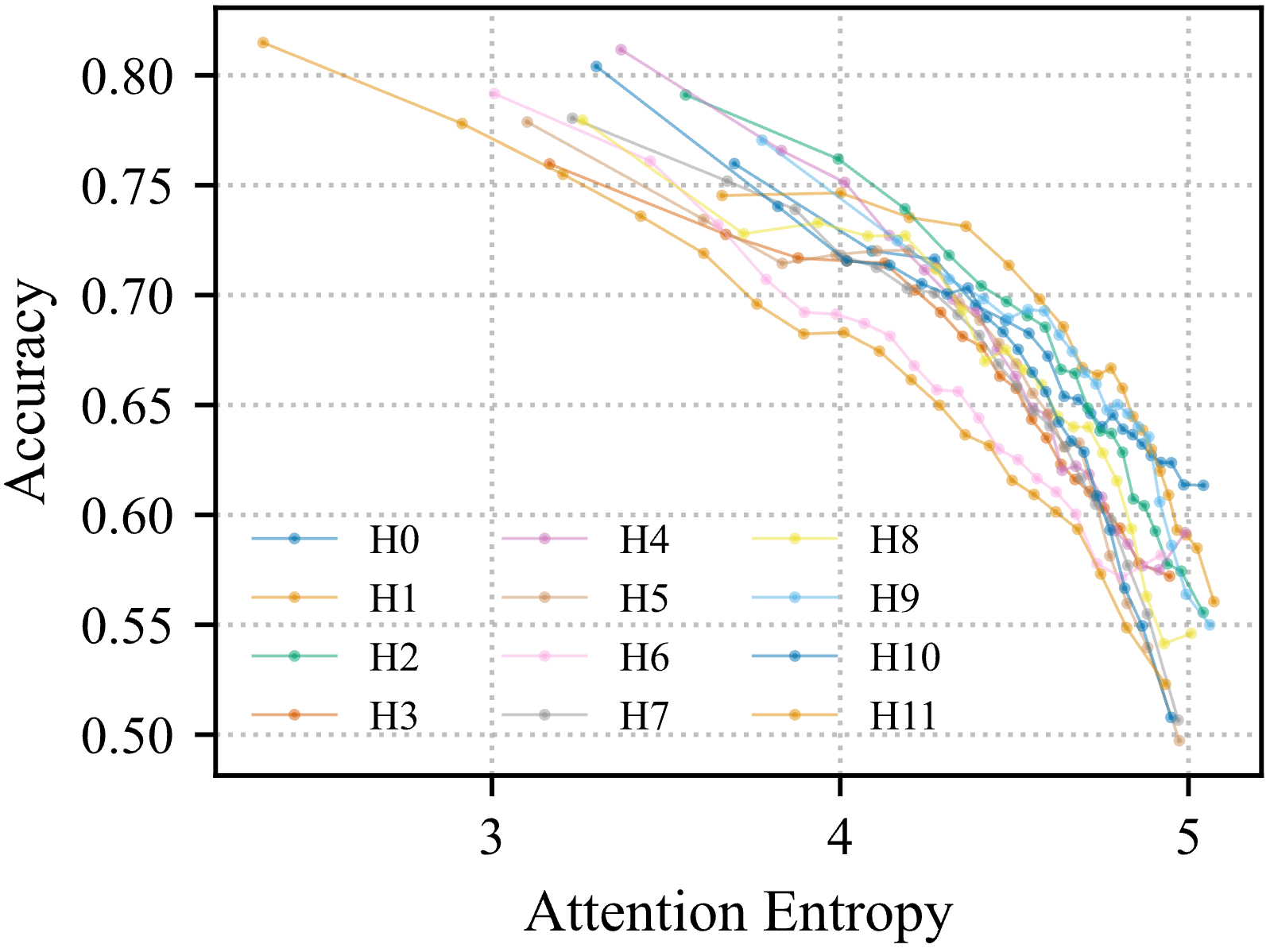}
  \caption{Attention Entropy vs Accuracy across Heads}
  \label{fig:correlation}
\end{subfigure}
\caption{%
\textbf{Left} (a) Our method: attention entropy minimization. 
We minimize the entropy of the attention distribution from CLS to patch tokens at test time.
We combine this with output entropy minimization for best results.
\textbf{Right}
(b) Visualization of attention entropy and accuracy.
The entropy of final-layer CLS to patch token attention and the accuracy on shifted data are shown across all heads on a 10\% sample of ImageNet-C for the unadapted DINOv3-Base model.
Higher entropy (x-axis, right) tends toward lower accuracy (y-axis, bottom).
}
\label{fig:combined_plots}
\end{figure}

We harness these attention distributions for TTA, minimizing the entropy of the attention distributions in vision transformers as an unsupervised loss to update the model parameters.
As this sharpens the distribution to focus more on fewer tokens, we call our method \textit{LookSharp}.
Specifically, we minimize the entropy of the distribution defined by the attention scores of the CLS token for the patch tokens from the last layer's attention heads.
Our approach is motivated by two key observations.
First, Figure~\ref{fig:combined_plots} (b) shows that accuracy drops sharply if the attention entropy is too diffuse.
Second, Modern ViTs like DINOv3 \citep{simeoni2025dinov3} learn interpretable and object-centric attention maps through internet-scale self-supervised training.

We demonstrate our method for adaptation to corruptions on ImageNet-C in the batch episodic setting.
That is, the model updates and then resets on each batch.
We also show that combining attention entropy and output entropy leads to further improvement.

% \section{Related Work}
% \label{gen_inst}

\textbf{Entropy Minimization for Adaptation.} Test-time adaptation often relies on entropy minimization.
Tent \citep{wang2021tent} updates normalization layer statistics and parameters to minimize output entropy.
MEMO \citep{NEURIPS2022_fc28053a} extends this by using test-time augmentation to create a batch from a single sample and updates all parameters episodically using the same loss as Tent.
Other works like SAR \citep{niu2023towards} and EATA (ETA) \citep{pmlr-v162-niu22a} use output entropy combined with sharpness-aware minimization, data filtering, and anchoring to the source model using regularization of the parameters.

\textbf{Attention for Adaptation.}
There has been less use of attention for updates.
\textbf{Attent} \citep{Kojima2023RobustifyingViT} aligns test-time attention statistics with stored source statistics.
Unlike Attent, our method is purely test-time and does not require storing source statistics. 
Instead, it relies on the confidence of attention during inference alone.
We therefore only compare with other fully test-time updates. %and source-free losses.
% \footnote{We limit our experimental comparisons to fully test-time and source-free methods for fair comparison.}

\section{Method: Attention Entropy Minimization}
\label{headings}

Given a model $f_\theta$ trained on a source distribution $\mathcal{D}_{source}$, we encounter a batch from a shifted distribution $\mathcal{D}_{shift}$ at test time. For each test batch, $\mathcal{B} = \{x_i\}_{i=1}^B$ where $x_i \sim \mathcal{D}_{shift}$, we aim to:

\begin{enumerate}
\item Adapt model parameters $\theta$ using an unsupervised objective $\mathcal{L}(x_i; \theta)$.
\item Generate prediction $\hat{y}_i = f_{\theta'}(x_i)$ using the adapted model.
\item Reset the model parameters to the original pretrained state (episodic). %Optionally, reset the optimizer state.
\end{enumerate}

\textbf{Loss: Attention Entropy Minimization.}
Let $\tA(x_i) \in \R^{H \times T \times T}$ denote the post-softmax attention tensor from the final transformer layer for input image $x_i$, where $H$ is the number of attention heads, $T$ is the sequence length (CLS, register, and patch tokens), and $t_{\text{cls}}$ is the index of the CLS token. Let $\sP$ denote the set of patch-token indices (excluding the CLS and register tokens), with $P = |\sP|$. We extract scores for the CLS token attending to the patch tokens and renormalize them to form a distribution $a^{(h)}(x_i)$:
\begin{equation}
a^{(h)}_j (x_i) = \frac{\etA_{h, t_{\text{cls}}, j}(x_i)}{\sum_{k \in \sP} \etA_{h, t_{\text{cls}}, k}(x_i)}, \quad \mathcal{L}_{Attention}(x_i) = -\frac{1}{H} \sum_{h=1}^{H} \sum_{j \in \sP} a^{(h)}_j(x_i) \log a^{(h)}_j(x_i)
\end{equation}

We \textbf{exclude} the CLS token to itself and to register tokens attention scores as we want to focus on the spatial patches of the image as opposed to global information. Minimizing this loss encourages each attention head to place concentrated (low-entropy) focus on a smaller subset of patch tokens, rather than distributing attention more diffusely. Averaging the distributions first, then taking their entropy, was tried but performed worse. This is reasonable as heads tend to specialize \citep{NEURIPS2021_652cf383}. We utilize the last layers attention scores as they are the most mature. 

We found that combining the standard output entropy minimization, as in \citep{wang2021tent}, with attention entropy minimization further improves performance. We use the standard output entropy minimization loss \citep{wang2021tent}:
\begin{equation}
\mathcal{L}_{Output}(x_i) = - \sum_{c \in \mathcal{C}} p_{\theta}(c \mid x_i) \log p_{\theta}(c \mid x_i)
\end{equation}
where $\mathcal{C}$ denotes the set of classes and $p_{\theta}(c \mid x_i)$ is the model's predicted class probability.

Thus, our loss is:
\begin{equation}
\mathcal{L}_{Combined}(x_i) = \mathcal{L}_{Attention}(x_i) + \mathcal{L}_{Output}(x_i)
\end{equation}

\section{Experiments and Results}

\begin{table}[t]
\begin{center}
  \begin{tabular}{lcccccc}
    \toprule
    Corruption & \shortstack{Source\\$\phantom{X}$} & \shortstack{Tent\\(Online)} & \shortstack{Tent\\(Episodic)} & \shortstack{Output\\$\phantom{X}$} & \shortstack{Attention\\(\textit{Ours})} & \textbf{\shortstack{Combined\\(\textit{Ours})}} \\
    \midrule
    Brightness          & 76.06 & 76.48 & 76.54 & 76.92 & 76.79 & \textcolor{red!70!black}{77.08} \\
    Contrast            & 51.96 & 55.68 & 55.01 & 58.72 & 56.72 & \textcolor{red!70!black}{58.95} \\
    Defocus Blur        & 42.87 & 44.61 & 45.43 & 46.87 & \textcolor{red!70!black}{48.26} & 47.47 \\
    Elastic Transform   & 33.94 & 40.57 & 37.51 & 42.35 & \textcolor{red!70!black}{47.64} & 44.15 \\
    Fog                 & 57.20 & 59.77 & 59.16 & 61.55 & \textcolor{red!70!black}{62.91} & 62.31 \\
    Frost               & 49.72 & 51.71 & 51.23 & 53.92 & \textcolor{red!70!black}{55.38} & 54.58 \\
    Gaussian Noise      & 33.15 & 36.23 & 36.37 & 40.78 & 32.87 & \textcolor{red!70!black}{40.95} \\
    Glass Blur          & 21.68 & 31.11 & 26.75 & 37.95 & 38.20 & \textcolor{red!70!black}{38.69} \\
    Impulse Noise       & 37.05 & 31.08 & 39.66 & 42.14 & 41.01 & \textcolor{red!70!black}{43.19} \\
    JPEG Compression    & 59.27 & 61.57 & 60.91 & 62.04 & 61.87 & \textcolor{red!70!black}{62.33} \\
    Motion Blur         & 48.49 & 50.93 & 50.27 & 53.15 & 52.85 & \textcolor{red!70!black}{53.48} \\
    Pixelate            & 63.27 & 65.34 & 64.88 & 66.57 & \textcolor{red!70!black}{67.42} & 66.91 \\
    Shot Noise          & 35.33 & 39.33 & 38.70 & 44.69 & 38.99 & \textcolor{red!70!black}{45.06} \\
    Snow                & 56.94 & 59.37 & 58.82 & 61.70 & 59.15 & \textcolor{red!70!black}{61.96} \\
    Zoom Blur           & 46.11 & 49.52 & 48.83 & 52.73 & \textcolor{red!70!black}{53.32} & 53.15 \\
    \midrule
    Mean            & \shortstack{47.54\\$\phantom{X}$} & \shortstack{50.22\\(+2.68)} & \shortstack{50.01\\(+2.47)} & \shortstack{53.47\\(+5.93)} & \shortstack{52.89\\(+5.35)} & \textbf{\shortstack{\textcolor{red!70!black}{54.02}\\(+6.48)}} \\
    \bottomrule
  \end{tabular}
\end{center}

\caption{Top-1 Accuracy (\%) on ImageNet-C level 5 corruptions. We report the source model and test-time adaptation variants: Tent (online/episodic), output entropy, attention entropy, and their combination (\textit{LookSharp}). All results use batch size 128. Attention, output and combined reset all the parameters.}
\label{tab:main_results}
\end{table}

We experiment with the standard benchmark for test-time adaptation applied to image classification, using a common architecture and a recent self-supervised backbone. We consider the batch-wise episodic test-time adaptation setting where the parameters are reset after each batch ~\cite{NEURIPS2022_fc28053a}, and also compare to an online (no resetting) method \citep{wang2021tent}.
%\subsection{Experiments}

\textbf{Dataset:} We evaluate on ImageNet-C \citep{hendrycks2019robustness}, which augments the standard ImageNet validation set with 15 different corruption types at 5 levels. We only evaluate on level 5, which is the most severe level of shift. We also perform TTA on clean data to ensure our method maintains performance without distribution shift. %\todo{do level 3 and include in appendix?}

\textbf{Model:} We use DINOv3-Base~\citep{simeoni2025dinov3}, pretrained on an internet-scale image dataset.
We train a linear classification head with this representation on the source data (ImageNet training split) using the standard cross-entropy loss (a.k.a. linear probing).
This yields 83.57\% top-1 accuracy on the validation set.
The images are preprocessed to the standard ImageNet size ($224 \times 224$) as in \cite{krizhevsky2012imagenet}.

\textbf{Evaluation Protocol:} For each corruption type, we report per-corruption accuracy and mean corruption accuracy at level 5.
We use a batch size of 128. 
The data is loaded in a randomized order for each shift, and as a result, each batch contains a mix of classes.
We optimize by Adam \citep{kingma2015adam} with learning rate $5 \times 10^{-5}$ for all methods except Tent.
For Tent, we use $10^{-3}$ in the episodic setting and $10^{-5}$ in the online setting.
These values are selected by a learning-rate sweep on the level 5 test set with mean accuracy as the metric.
We perform $1$ gradient update per batch and update all parameters.

\textbf{Baselines:} We evaluate without any test-time updates, to measure the robustness of the source model. We also compare with Tent \citep{wang2021tent}, where only the normalization layer parameters are updated, in the episodic and in the online case.

% \section{Results}

\textbf{Results.} Table~\ref{tab:main_results} shows that our method improves mean accuracy compared to the non-adapted source model on ImageNet-C. The output-head entropy loss alone performs better than attention entropy alone, but combining both losses yields even better results. On clean data, the attention-only loss \textit{slightly} hurts performance ($83.57\% \rightarrow 82.95\%$). Using the combined loss \textit{slightly} improves accuracy ($83.57\% \rightarrow 83.80\%$).

\begin{figure}[t]
\centering
\includegraphics[width=\linewidth]{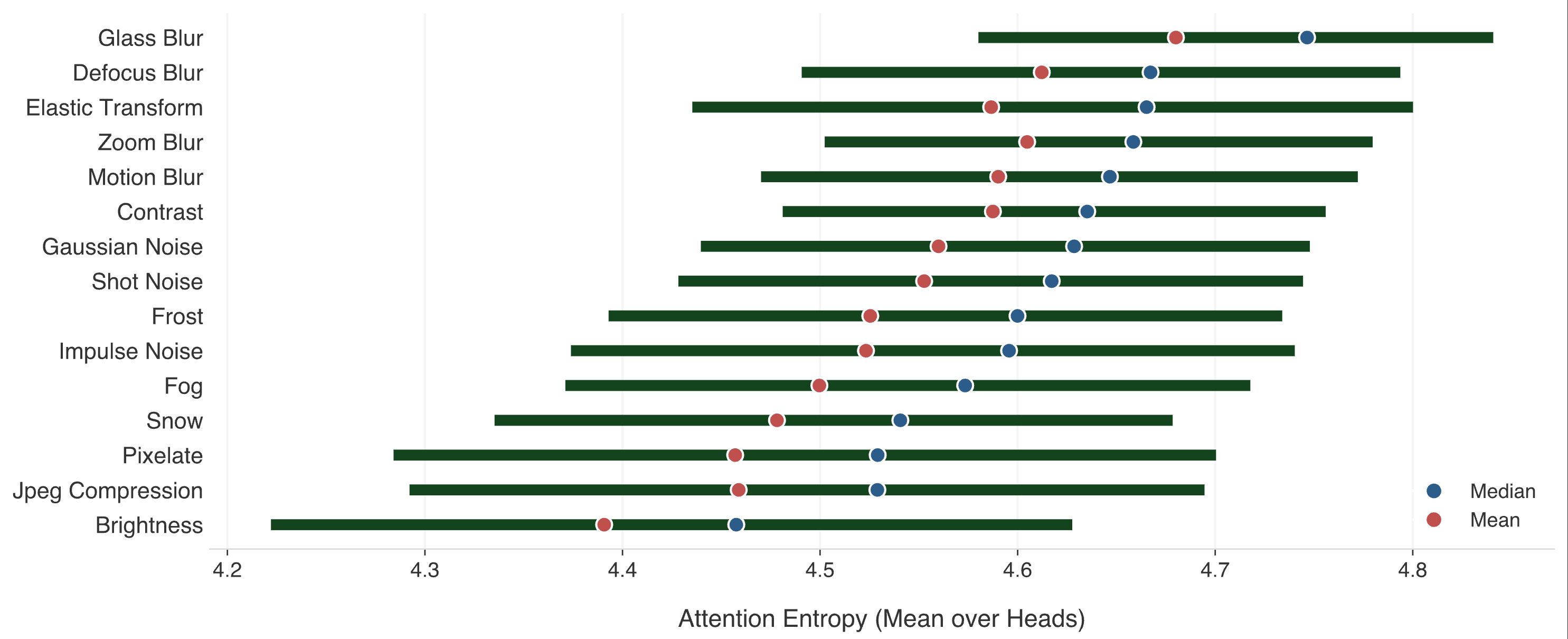}
\caption{
  Median and interquartile range (IQR) of per-image mean attention entropy across a 10\% sample of ImageNet-C at level 5. For each corruption, attention entropy is first averaged over heads for each image. Blurs and blur-like corruptions tend to have higher attention entropies.
}
\label{fig:attentpercor}
\end{figure}

Overall, our combined objective achieves the best mean corruption accuracy, improving mean accuracy from 47.54\% (Source) to 54.02\% (+6.48 \%). Attention-based entropy minimization works best for blur and blur-like corruptions (elastic transform). We can see from Figure~\ref{fig:attentpercor} that this is because blurring images makes the attention maps more diffuse, and this is what $\mathcal{L}_{Attention}$ is directly adressing. A visualization of the attention loss is shown in Appendix~\ref{sec:appendix}.

In our experiments, we found that Tent (Online) is highly sensitive to the learning rate, consistent with \citet{pmlr-v202-zhao23d}. Larger learning rates improve performance on some corruptions but cause the model to collapse on others, resulting in mean accuracy below the source model. The learning rate we select is the one that achieves maximal mean accuracy on the level 5 test set.

\section{Conclusion and Future Work}

We introduce \textit{LookSharp}, a simple test-time adaptation method that minimizes the entropy of CLS-to-patch attention, and show consistent gains on ImageNet-C, especially for blur-like corruptions.
Combining attention and output entropy yields the best overall accuracy, suggesting the two signals are complementary.

\textbf{Limitations.} The method incurs computational overhead due to the forward-backward-forward passes needed and requires self-attention in the model architecture. Attention-based adaptation likely also depends on the quality of the learned attention maps, which vary across architectures and pretraining regimes \citep{darcet2024visiontransformersneedregisters}.

% \textbf{Future work.} Extending this approach to other fields like natural language processing (NLP), where sequence lengths are much longer, would require some modifications but has potential for domain adaptation in NLP.

While this work focuses on succinct experiments to show the effectiveness of attention entropy as an unsupervised TTA loss, future work can explore trying to extract more performance by exploring a dynamic weighting of attention and output entropy based on input characteristics or multi-layer attention losses that span the model from shallow to deep.

\section*{Acknowledgements}
We thank Vivian White for helpful discussions. We also thank the Digital Research Alliance of Canada and the Vector Institute for computational resources.

%and investigate the theoretical link between optimizer moments and heavy-tailed attention distributions.

% \subsubsection*{Acknowledgments}
% Use unnumbered third level headings for the acknowledgments. All
% acknowledgments, including those to funding agencies, go at the end of the paper.

\begingroup
\setlength{\emergencystretch}{6em}
\sloppy
\raggedright
\bibliographystyle{iclr2026_conference}
\bibliography{iclr2026_conference}
\endgroup

\clearpage

\appendix

\section{Appendix}
\label{sec:appendix} % <-- Add this label here

The figure below shows the CLS-to-patch attention distribution of the model taken from the final layer and averaged across the heads. 

\begin{figure}[h]
    \centering
    \includegraphics[width=1\linewidth]{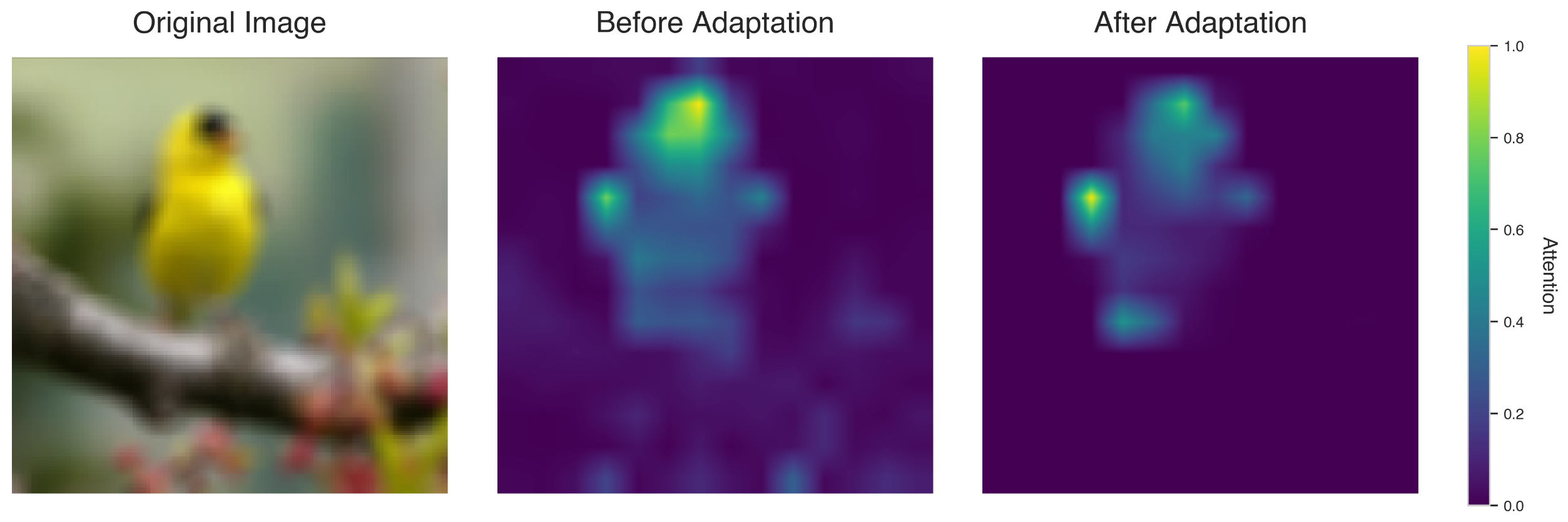}
    
    \includegraphics[width=1\linewidth]{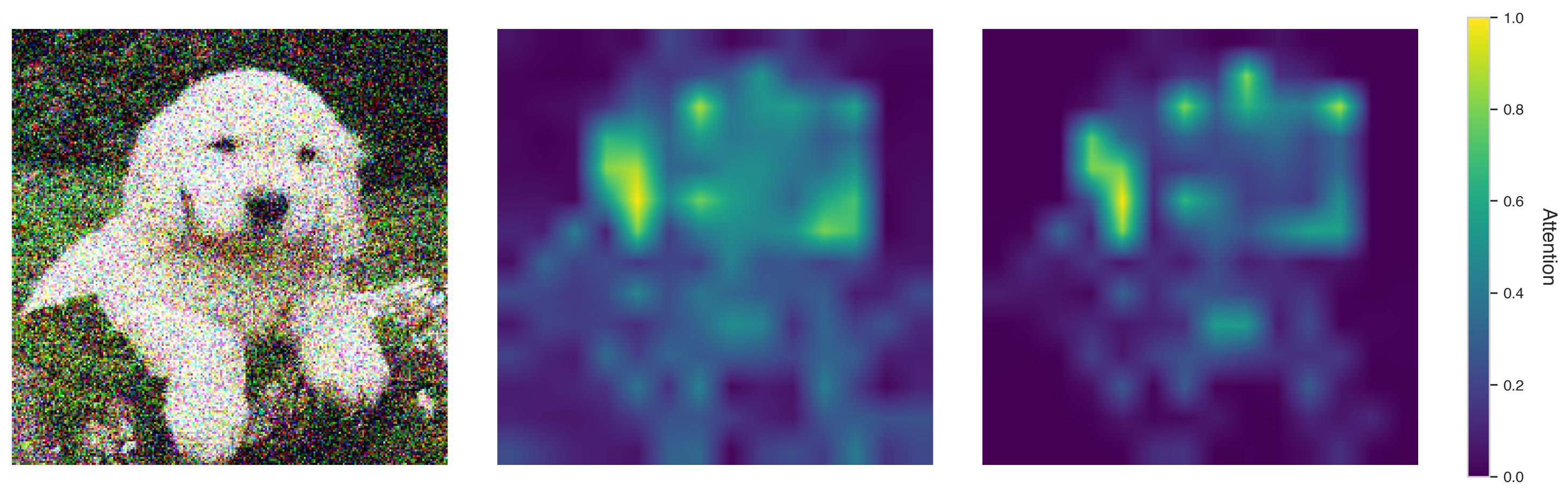}
    
    \caption{This shows the attention map before and after adaptation using our attention entropy loss $(\mathcal{L}_{Attention})$.}
    \label{fig:attnmap}
\end{figure}

% \subsection{Examples}

% \todo{Before and after attention "spread" on 2-3 images}

\end{document}

%% file: math_commands.tex
\usepackage{amsmath,amsfonts,bm}

\def\eqref#1{equation~\ref{#1}}
\def\1{\bm{1}}

\DeclareMathAlphabet{\mathsfit}{\encodingdefault}{\sfdefault}{m}{sl}
\SetMathAlphabet{\mathsfit}{bold}{\encodingdefault}{\sfdefault}{bx}{n}
\newcommand{\tens}[1]{\bm{\mathsfit{#1}}}
\def\tA{{\tens{A}}}

\def\sP{{\mathbb{P}}}

\newcommand{\etens}[1]{\mathsfit{#1}}

\def\etA{{\etens{A}}}

\newcommand{\R}{\mathbb{R}}

%% file: iclr2026_conference.bib
@article{hendrycks2019robustness,
  title={Benchmarking Neural Network Robustness to Common Corruptions and Perturbations},
  author={Dan Hendrycks and Thomas Dietterich},
  journal={Proceedings of the International Conference on Learning Representations},
  year={2019}
}

@misc{dosovitskiy2021imageworth16x16words,
      title={An Image is Worth 16x16 Words: Transformers for Image Recognition at Scale}, 
      author={Alexey Dosovitskiy and Lucas Beyer and Alexander Kolesnikov and Dirk Weissenborn and Xiaohua Zhai and Thomas Unterthiner and Mostafa Dehghani and Matthias Minderer and Georg Heigold and Sylvain Gelly and Jakob Uszkoreit and Neil Houlsby},
      year={2021},
      eprint={2010.11929},
      archivePrefix={arXiv},
      primaryClass={cs.CV},
      url={https://arxiv.org/abs/2010.11929}
}

@misc{simeoni2025dinov3,
  title={{DINOv3}},
  author={Sim{\'e}oni, Oriane and Vo, Huy V. and Seitzer, Maximilian and Baldassarre, Federico and Oquab, Maxime and Jose, Cijo and Khalidov, Vasil and Szafraniec, Marc and Yi, Seungeun and Ramamonjisoa, Micha{\"e}l and Massa, Francisco and Haziza, Daniel and Wehrstedt, Luca and Wang, Jianyuan and Darcet, Timoth{\'e}e and Moutakanni, Th{\'e}o and Sentana, Leonel and Roberts, Claire and Vedaldi, Andrea and Tolan, Jamie and Brandt, John and Couprie, Camille and Mairal, Julien and J{\'e}gou, Herv{\'e} and Labatut, Patrick and Bojanowski, Piotr},
  year={2025},
  eprint={2508.10104},
  archivePrefix={arXiv},
  primaryClass={cs.CV},
  url={https://arxiv.org/abs/2508.10104}
}

@inproceedings{
wang2021tent,
title={Tent: Fully Test-Time Adaptation by Entropy Minimization},
author={Dequan Wang and Evan Shelhamer and Shaoteng Liu and Bruno Olshausen and Trevor Darrell},
booktitle={International Conference on Learning Representations},
year={2021},
url={https://openreview.net/forum?id=uXl3bZLkr3c}
}

@inproceedings{NEURIPS2022_fc28053a,
 author = {Zhang, Marvin and Levine, Sergey and Finn, Chelsea},
 booktitle = {Advances in Neural Information Processing Systems},
 editor = {S. Koyejo and S. Mohamed and A. Agarwal and D. Belgrave and K. Cho and A. Oh},
 pages = {38629--38642},
 publisher = {Curran Associates, Inc.},
 title = {MEMO: Test Time Robustness via Adaptation and Augmentation},
 url = {https://proceedings.neurips.cc/paper_files/paper/2022/file/fc28053a08f59fccb48b11f2e31e81c7-Paper-Conference.pdf},
 volume = {35},
 year = {2022}
}

@inproceedings{darcet2024visiontransformersneedregisters,
  title     = {Vision Transformers Need Registers},
  author    = {Darcet, Timoth{\'e}e and Oquab, Maxime and Mairal, Julien and Bojanowski, Piotr},
  booktitle = {Proceedings of the International Conference on Learning Representations (ICLR)},
  year      = {2024}
}

@inproceedings{kingma2015adam,
  title     = {Adam: A Method for Stochastic Optimization},
  author    = {Kingma, Diederik P. and Ba, Jimmy},
  booktitle = {International Conference on Learning Representations (ICLR)},
  year      = {2015}
}

@inproceedings{NEURIPS2021_652cf383,
 author = {Raghu, Maithra and Unterthiner, Thomas and Kornblith, Simon and Zhang, Chiyuan and Dosovitskiy, Alexey},
 booktitle = {Advances in Neural Information Processing Systems},
 editor = {M. Ranzato and A. Beygelzimer and Y. Dauphin and P.S. Liang and J. Wortman Vaughan},
 pages = {12116--12128},
 publisher = {Curran Associates, Inc.},
 title = {Do Vision Transformers See Like Convolutional Neural Networks?},
 url = {https://proceedings.neurips.cc/paper_files/paper/2021/file/652cf38361a209088302ba2b8b7f51e0-Paper.pdf},
 volume = {34},
 year = {2021}
}

@article{Kojima2023RobustifyingViT,
  title     = {Robustifying Vision Transformer Without Retraining from Scratch Using Attention-Based Test-Time Adaptation},
  author    = {Kojima, Takuya and Iwasawa, Yusuke and Matsuo, Yutaka},
  journal   = {New Generation Computing},
  volume    = {41},
  pages     = {5--24},
  year      = {2023},
  doi       = {10.1007/s00354-022-00197-9},
  publisher = {Springer},
  received  = {2022-09-02},
  accepted  = {2022-11-24},
  published = {2022-12-27}
}

@InProceedings{pmlr-v162-niu22a,
  title = 	 {Efficient Test-Time Model Adaptation without Forgetting},
  author =       {Niu, Shuaicheng and Wu, Jiaxiang and Zhang, Yifan and Chen, Yaofo and Zheng, Shijian and Zhao, Peilin and Tan, Mingkui},
  booktitle = 	 {Proceedings of the 39th International Conference on Machine Learning},
  pages = 	 {16888--16905},
  year = 	 {2022},
  editor = 	 {Chaudhuri, Kamalika and Jegelka, Stefanie and Song, Le and Szepesvari, Csaba and Niu, Gang and Sabato, Sivan},
  volume = 	 {162},
  series = 	 {Proceedings of Machine Learning Research},
  month = 	 {17--23 Jul},
  publisher =    {PMLR},
  url = 	 {https://proceedings.mlr.press/v162/niu22a.html}
}

@InProceedings{pmlr-v202-zhao23d,
  title = 	 {On Pitfalls of Test-Time Adaptation},
  author =       {Zhao, Hao and Liu, Yuejiang and Alahi, Alexandre and Lin, Tao},
  booktitle = 	 {Proceedings of the 40th International Conference on Machine Learning},
  pages = 	 {42058--42080},
  year = 	 {2023},
  editor = 	 {Krause, Andreas and Brunskill, Emma and Cho, Kyunghyun and Engelhardt, Barbara and Sabato, Sivan and Scarlett, Jonathan},
  volume = 	 {202},
  series = 	 {Proceedings of Machine Learning Research},
  month = 	 {23--29 Jul},
  publisher =    {PMLR},
  url = 	 {https://proceedings.mlr.press/v202/zhao23d.html}
}

@inproceedings{krizhevsky2012imagenet,
  title     = {ImageNet Classification with Deep Convolutional Neural Networks},
  author    = {Krizhevsky, Alex and Sutskever, Ilya and Hinton, Geoffrey E.},
  booktitle = {Advances in Neural Information Processing Systems},
  year      = {2012}
}

@inproceedings{niu2023towards,
  title={Towards Stable Test-Time Adaptation in Dynamic Wild World},
  author={Niu, Shuaicheng and Wu, Jiaxiang and Zhang, Yifan and Wen, Zhiquan and Chen, Yaofo and Zhao, Peilin and Tan, Mingkui},
  booktitle = {Internetional Conference on Learning Representations},
  year = {2023}
}

@misc{fullerlookwhere,
      title={LookWhere? Efficient Visual Recognition by Learning Where to Look and What to See from Self-Supervision}, 
      author={Anthony Fuller and Yousef Yassin and Junfeng Wen and Daniel G. Kyrollos and Tarek Ibrahim and James R. Green and Evan Shelhamer},
      year={2025},
      eprint={2505.18051},
      archivePrefix={arXiv},
      primaryClass={cs.CV},
      url={https://arxiv.org/abs/2505.18051}, 
}
